\documentclass[a4wide]{article}

\usepackage{rotating}
\usepackage{pdflscape}

\usepackage[numbers,sort&compress,longnamesfirst,sectionbib]{natbib}

\usepackage[labelfont=bf,font=footnotesize]{caption}

\usepackage{amsmath}

\usepackage[percent]{overpic}

\usepackage{xspace}
\usepackage{balance}

\usepackage{enumitem}

\usepackage[margin=1.5in]{geometry}
\usepackage{helvet}  
\usepackage{courier}  
\usepackage{url}  
\usepackage{graphicx,dsfont}  

\usepackage{ragged2e}
\usepackage{float}
\usepackage{booktabs} 
\usepackage{amsmath}
\usepackage{graphicx}
\usepackage{algorithmic,algorithm}
\usepackage{pdfpages}
\usepackage{subfig}

\newcommand{\mlea}{(\mu + \lambda)-EA_D}

\newcommand{\GCF}{\mbox{\em{GCF}}}

\newcommand{\symm}{\mbox{\em{Symm}}}
\newcommand{\hue}{\mbox{\em{Hue}}}
\newcommand{\sat}{\mbox{\em{Saturation}}}

\newcommand{\smoothl}{\mbox{\em{smoothness}}}

\newcommand{\gan}{\mbox{\em{GAN}}}

\newcommand{\cma}{\mbox{\em{CMA-ES}}}
\newcommand{\EA}{\mbox{\em{(1+1) EA}}}
\newcommand{\f}{\mbox{\em{feature}}}
\newcommand{\fone}{\mbox{\em{feature 1}}}
\newcommand{\ftwo}{\mbox{\em{feature 2}}}
\newcommand{\discr}   {  {|} \mbox{\em{discriminator}}{|}}

\title{Evolution of Images with Diversity and Constraints Using a Generator Network}

\author{
Aneta Neumann\\
Optimisation and Logistics\\
School of Computer Science\\
The University of Adelaide\\
Adelaide, Australia\\
{aneta.neumann@adelaide.edu.au}
\and 
Christo Pyromallis\\
School of Computer Science\\
The University of Adelaide\\
Adelaide, Australia\\
{christo.pyromallis@}\\{student.adelaide.edu.au}\\
\and
Bradley Alexander\\
Optimisation and Logistics\\
School of Computer Science\\
The University of Adelaide\\
Adelaide, Australia\\
{bradley.alexander@adelaide.edu.au}
}

\begin{document}

\maketitle

\begin{abstract} 

Evolutionary search has been extensively used to generate artistic images. Raw images have high dimensionality which makes a direct search for an image challenging. In previous work this problem has been addressed by using compact symbolic encodings or by constraining images with priors. Recent developments in deep learning have enabled a generation of compelling artistic images using generative networks that encode images with lower-dimensional latent spaces. 
To date this work has focused on the generation of images concordant with one or more classes and transfer of artistic styles. There is currently no work which uses search in this latent space to generate images scoring high or low aesthetic measures.
In this paper we use evolutionary methods to search for images in two datasets, faces and butterflies, and demonstrate the effect of optimising aesthetic feature scores in one or two dimensions. The work gives a preliminary indication of which feature measures promote the most interesting images and how some of these measures interact.

\end{abstract}

\section{Introduction}
\label{introduction}
Evolutionary search has frequently been used to generate artistic  images~\cite{Heijer14,correia2013machado,DBLP:journals/tec/KowaliwDM12}. Images have a high-dimension. Previous work has either reduced the dimensionality of the search space through programmatic encodings~\cite{correia2013machado,citeulike:12541313} or have constrained the images with priors~\cite{neumann2017evolutionary,DBLP:conf/gecco/AlexanderKN17}. In recent years, Generative Adversarial Networks\cite{goodfellow2014generative} \gan's have been used to map a low dimensional real-valued latent vector into images in the category for which the ${\gan}$ has trained. There has been work in using \gan's to generate and mix novel images~\cite{nguyen2016a} and perform artistic style transfer~\cite{gatys2016image}, along with many other applications. However, to date there has no work using evolutionary, or other methods, to explore the latent space of a ${\gan}$ to generate images according to aesthetic feature measures. In this work we employ \gan's to generate novel images by evolving the latent vector to maximise and minimise single aesthetic and pairs of aesthetic features. We show that the generation of images in this space requires the use of carefully constructed constraints on image realism. We also show that different \gan's appear to impose different bounds on the values of aesthetic measures that can be evolved. 

The paper is structured as follow. Section \ref{related} outlines related work. Section \ref{basic} presents the methodology used for evolving images. Section \ref{preliminary} and \ref{results} present our results. Finally, section \ref{discussion} presents discussion and the future work.

\subsection{Related Work}
\label{related}
Aesthetic feature measures have been often applied to the creation of new artistic images using evolutionary search~\cite{Heijer14,correia2013machado,DBLP:journals/tec/KowaliwDM12,machado2008experiments}. 
There has also been significant work in the evolution of existing images~\cite{DBLP:conf/gecco/NeumannSCN17,DBLP:conf/iconip/NeumannAN16}.

This work differs from previous work in the use of a ${\gan}$ as a mapper from the latent search vector to the images space and the use of the discriminator network and feature metrics to constrain these images, 
the space of a generator network as a given image source and use the discriminator and feature metrics to constrain these images.
In terms of deep learning, 
 Gatys~\cite{gatys2016image,DBLP:conf/cvpr/GatysEBHS17,DBLP:conf/nips/GatysEB15} used a convolutional network to transfer artistic style into an existing image.  These new approaches in network architectures and training
methods enabled the generation of  realistic images~\cite{radford2015unsupervised,dosovitskiy2015learning}.
Recently Dosovitskiy and Brox~\cite{dosovitskiy2016generating} trained networks of generating images from feature vector and 
combining an auto-encoder-style approach with deep convolutional generative adversarial networks training. Furthermore, Nguyen~\cite{nguyen2016a} used priors from Deep Generative Network to generate image variants that look close to natural within a preferred inputs for neurons.
 
For our investigation we consider recently work~\cite{DBLP:conf/gecco/AlexanderKN17} on feature based diversity optimization. The approach applied $\mlea$ to evolve diverse image instances. 
Previously the algorithm was introduced in~\cite{gao2016feature} for Traveling Salesman Problem (TSP) which is a NP-hard combinatorial optimization problem with real world applications.

\begin{figure} [!th]
\centering
\includegraphics[width=7cm]{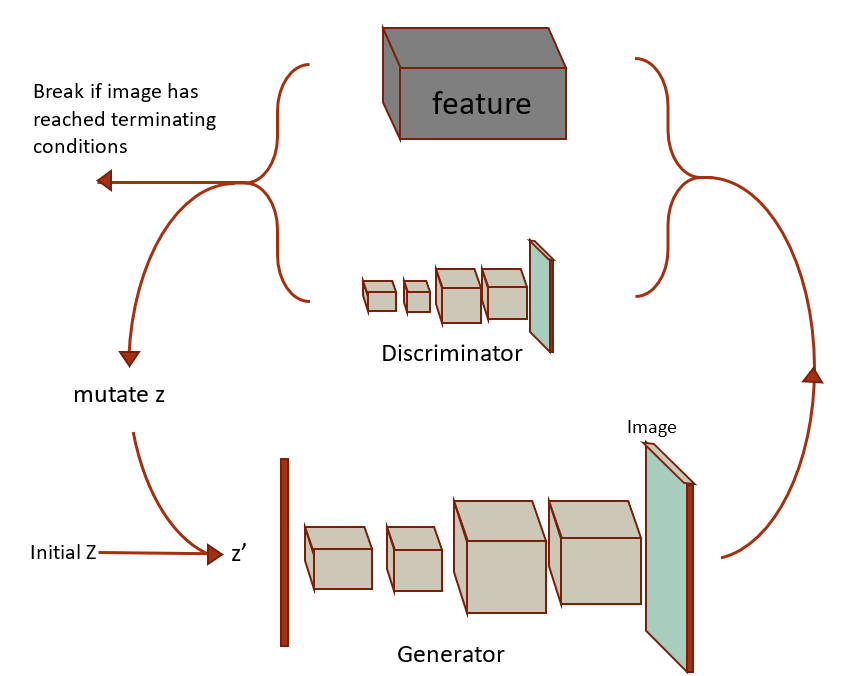}

\caption{The final setup of the system. The latent vector $Z$ is randomly seeded and sent through the system, mutating until it reaches an optimal solution or the termination condition ($2000$ mutations) is reached.}
\label{model_1}
\end{figure}

\section{Methodology}
\label{basic}
In this section we describe the methodology that we use to evolve images. In our system, images are created by optimising the latent space of the ${\gan}$ to create images that score high or low on aesthetic feature measures. Further, we try to constrain images to be real, this is done using the discriminator network trained with the ${\gan}$.

\subsection{Our System}
In this section we describe our system that is based on Generative Adversarial Networks ${\gan}$~\cite{goodfellow2014generative}. Figure~\ref{model_1} shows the structure of our system. In the ${\gan}$, we train two networks: 1) a generator, to generate images from a latent vector $z$ of 100 real numbers,  and 2) a discriminator which scores the images in the generator for realness as both networks are trained. We train these two networks with two image datasets: the Celebfaces attributes ({\em{CelebA}}) data set containing more than 150k celebrity faces~\cite{liu2015faceattributes} and the imagenet~\cite{deng2009imagenet} class of butterflies containing over $45000$ butterfly images. 

The ${\gan}$ is based on the Pytorch ${\gan}$ tutorial~\cite{pytorchGan2017}. 
The generator component of the network consists of $5$ deconvolutional layers. The activation functions for the first $4$ layers are LeakyRelu's, and the hidden layer is a ReLU~\cite{nair2010rectified}, and the last deconvolutional layer uses {\em{tanh}}. The discriminator uses LeakyReLU's in its $5$ convolutional layers.  
The generator takes the hundred elements of $Z$ as input and generates a 128x128 pixel image. The discriminator takes an image and generates a normally distributed {\em{realness}} score with the most real images scoring zero. 
Finally, at the top of Figure~\ref{model_1} is the aesthetic feature 
function, as described in~\cite{DBLP:conf/gecco/AlexanderKN17}.

The ${\gan}$ and the necessary feature functions are linked together to drive evolution. The combined system works as follows. A randomly initialised latent feature vector is sent into the generator, which outputs an image. This image is run through both the chosen artistic feature function and the discriminator, and both contribute to a score. The evolutionary process, guided by the score, mutates $Z$ with the goal of optimising both the realness and the desired artistic feature of the output image.

\subsection{Aesthetic Features}
This section describes in more detail the features used in our experiments. We denote a function $f$ for an image $I$, representing an artistic feature. This function maps an image $I$ to a scalar value $f(I)$. For our experiments we use the following features: mean hue, saturation, smoothness, reflectional symmetry and Global Contrast Factor. We describe the features as follows.

{\em{Mean-hue}} is the mean value of every pixel's hue in the image. The range of $\hue$ is $[0,1]$. Note that both $0$ and $1$ represent the colour red. 

{\em{Mean-saturation}} is the mean value of every pixel's saturation in the image. The range is $[0,1]$ with $0$ representing low $\sat$ and $1$ representing high.

{\em{Smoothness}} of an image is measured, for a given picture $I$ with $N$ pixels as:
\[
1- \sum\nolimits_{i=1}^{N} \sum\nolimits_{c=1}^{3} \mbox{\em{gradient}}(I_{ic})/3N,
\]

where {\em{gradient}} is the gradient magnitude image produced by MATLAB's {\em{intermediate}} image gradient method, which calculated gradients between adjoining pixel values on each colour channel. We perceive from this that $\smoothl(I)$ is the disparity of colour between adjacent pixels and also lies within the range $[0,1]$. 

\emph{Reflectional Symmetry} is a measure based on den Heijer's work~\cite{Heijer14} to measure the degree which an image reflects itself. Symmetry divides an image into four quadrants and measures horizontal, vertical, and diagonal symmetry. Note 
Symmetry is defined for image $I$ as:
\[
\symm(I) = S_h(I)+S_v(I)+S_d(I)/3
\]

{\emph{Global Contrast Factor}} ${(\GCF)}$ is a measure of mean contrast between neighbouring pixels at different image resolutions. ${\GCF}$ is determined by calculating the local contrast at each pixel at resolution $r$: 
\[
lc_r(I_{ij})=\sum\nolimits_{I_{kl} \in N(I_{ij})} |lum (I_{kl}) -lum(I_{ij})|
\]
where $lum(P)$ is the perceptual luminosity of pixel $P$ and $N(I_{ij})$ are the four neighbouring pixels of $I_{ij}$ at resolution $r$. The mean local contrast at the current resolution is defined as: 
\[
C_r=(\sum\nolimits_{i=1}^m \sum\nolimits_{j=1}^n lc_r(I_{ij}))/(mn).\] From these local contrasts, ${\GCF}$ is calculated as
\[
\GCF = \sum\nolimits_{r=1}^9 w_r \cdot C_r.
\]
The pixel resolutions correspond to different {\em{superpixel}} sizes of $1,2,4,8,16,25,50,100$, and $200$. Each superpixel is set to the average luminosity of the pixel's it contains. The $w_r$ are empirically derived weights of resolutions from~\cite{matkovic2005global} giving highest weight to moderate resolutions. Note ${\GCF}$'s range is not bounded to $[0,1]$.

\subsection{Feature Optimisation}

In this work we investigated the use of single and multi-feature optimization. We explore optimization space with respect to feature values. For a single feature our system minimise and maximise particularly feature. We define the minimization process ${\f}$ and for maximization process $(1.0 - ${\f}$)$\footnote{Note that for feature${\GCF}$ we are able to maximize $1/{\GCF}$ and scale is in the range $[0,1]$.}. For double features $(f,g)$ we have four optimisaton targets representing the combinations of minimising and maximising $f$ and $g$. 
 
 To maintain realness we penalise the combined asthetic score with a measure for realness from the discriminator $\discr$.
 Thus our fitness functions for single features are shown in equations~
$(1-2)$ and for multi-features in $(3-6)$.

\begin{gather}
\f \times \discr \\
(1.0 - \f ) \times \discr \\
\fone \times \ftwo \times \discr \\
\fone  \times (1.0 - \ftwo ) \times \discr \\
(1.0 - \fone) \times \ftwo \times \discr \\
(1.0 - \fone) \times (1.0 - \ftwo ) \times \discr
\end{gather}

For multi-feature experiments we use $6$ feature combinations: hue-saturation, hue-symmetry, saturation-symmetry, smoothness-saturation, ${\GCF}$-smoothness and ${\GCF}$-saturation.
These combinations were chosen to produce potentially interesting outputs. ${\GCF}$ + smoothness and ${\GCF}$ + saturation were specifically chosen due to related work indicating ${\GCF}$ + smoothness would constrain each other ~\cite{DBLP:conf/gecco/AlexanderKN17}, resulting in lower image diversity.

\section{Preliminary Experiments}
\label{preliminary}
To ensure meaningful  results, we refined the methodology through an experimental process. These refinements are discussed in following.

Initial experimentation was done using ${\EA}$ and ${\cma}$~\cite{hansen2003reducing} frameworks to determine the performance of both algorithms. ${\cma}$ was allowed to run for $2000$ mutations (the equivalent of $80000$ iterations) and ${\EA}$ was allowed to run for $80000$ iterations to promote a valid comparison. 

\begin{figure}[th]
\centering
\includegraphics[width=8.12cm]{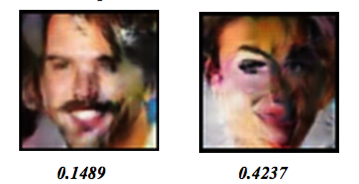}
\caption{Image obtained by using evolutionary algorithms: ${\EA}$ and ${\cma}$, with corresponding hue feature value, from left respectively.}
\label{1+1_EA_CMS_ES_2}
\end{figure}

As illustrated in Figure~\ref{1+1_EA_CMS_ES_2} $\cma$ was able to achieve more extreme feature values. This superiority of optimisation applied to all feature metrics. The ${\cma}$ ran for 80 minutes and ${\EA}$ taking 240 minutes for an optimisation run.  

\begin{figure}
\centering


\subfloat[Hue] {
\includegraphics[width=3.1cm]{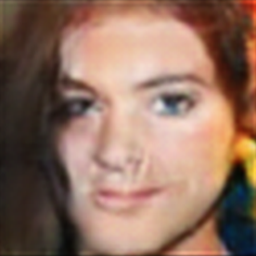}
\includegraphics[width=3.1cm]{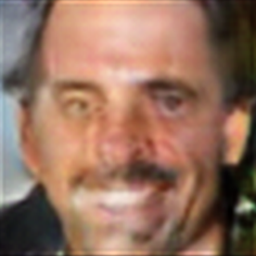}}
\subfloat[Symmetry] {
\includegraphics[width=3.1cm]{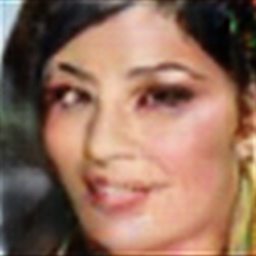}
\includegraphics[width=3.1cm]{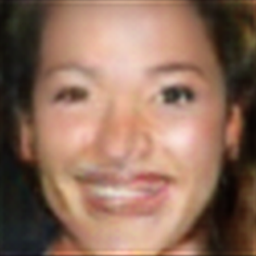}}    

\caption{Image obtained by single features. The left column corresponds to minimizing features hue and symmetry, from top with value 0.0841 and 0.7963, respectively. The right column corresponds to maximizing features hue and symmetry, from top with value 0.1275 and 0.9198, respectively. }
\label{single_feature_hue_symmetry_4}
\end{figure}

\subsection{Feature Experiments without Realness Constraint}
\begin{figure}[th]
\centering
\includegraphics[width=8.12cm]{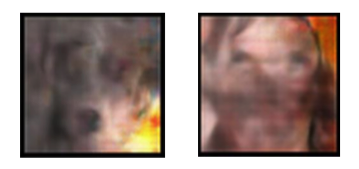}
\caption{Image obtained without realness constraints by minimizing saturation and hue, from links respectively.}
\label{without_realness_sat_hue_3}
\end{figure}

It was initially assumed that the ${\gan}$ would be able to create face-like images with any input vector. Some tests were performed which did not incorporate a constraint on realness as part of the optimisation process. As Figure~\ref{without_realness_sat_hue_3} demonstrates optimising features without the discriminator produces abstract images.  
To constrain images to be more realistic three constraining methods were tested: 1) limiting the degree ${\cma}$ was allowed to mutate $Z$; 2) discarding images that failed a certain realness threshold and 3) incorporating the discriminator's return value into the optimisation function. It was found that the third option of integrating realness as a variable $\discr$  gave the best results. The option of 
discarding images resulted in $\cma$ failing progress. Restricting the values in $Z$ to near-zero do not preserve realism as much as using the discriminator.

\subsection{Single Dimensional Feature Experiments with Constraint}
Single feature experiments require the fewest variables to optimise and as such, could be expected to evolve the image with the least difficulty. The results of each feature value are shown in  Table~\ref{tab:single}.

\begin{table}
\centering
\caption{Single dimensional feature values obtained from experiments with constraint.}\label{tab:single}
\begin{tabular}{l|r|r}
\hline
Feature  & Min & Max\\\hline
$\hue$ & 0.0841 & 0.1275\\\hline
$Saturation$ & 0.3306 & 0.3543  \\\hline 
$Smoothness$ & 0.9737 & 0.9843\\\hline
$Symmerty$ & 0.7985  & 0.9198\\\hline
$\GCF$ &0.0276&0.0286 \\\hline 
\end{tabular}
\end{table}
As can be seen the ranges of features above are very small, with the exception of symmetry. In these runs the $\discr$ term has the strong effect of constraining the feature values. For symmetry, the larger range might be explained by presence of both symmetric and asymmetric faces in the training dataset. 
In line with the small feature ranges the constrained 
images only showed small differences as seen in figure~\ref{single_feature_hue_symmetry_4} corresponding to 
the hue and symmetry measures in Table~\ref{tab:single}.

\subsection{Two-Dimensional Feature Experiments with Constraint}
\begin{figure}

\centering
\includegraphics[width=3.1cm]{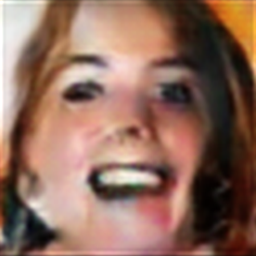}
\vspace{0.08cm}
\includegraphics[width=3.1cm]{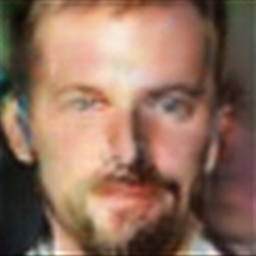}\\
\includegraphics[width=3.1cm]{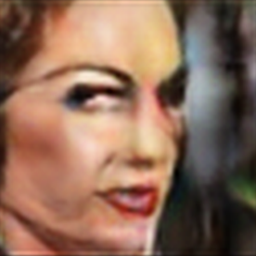}
\includegraphics[width=3.1cm]{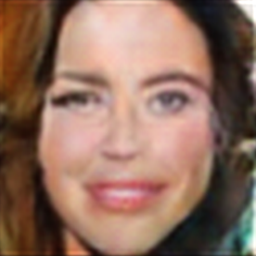}

\caption{ Images obtained by multi-features. The left column corresponds to minimizing feature saturation and symmetry, from top, respectively. The right column corresponds to maximizing feature saturation and symmetry, from too, respecively.}
\label{multi_features_saturat_symm_5}
\end{figure}

Running the experiments on multiple features gave similarly constrained images to the single features. Figure~\ref{multi_features_saturat_symm_5} shows images evolved to minimise and maximise saturation and symmetry.

As can be seen in the Figure~\ref{multi_features_saturat_symm_5}, there is some success in evolving different amounts of symmetry but not a particularly strong difference in saturation. It appears, at least for some features, the realness constraint is preventing strong exploration of the feature space.

\subsection{Impact of cut-off function}
In order to maintain balance between 
image realness and exploration we modify the discriminator term by passing the raw result of $\discr=x$ for an image through a cut-off function $f$ define as follow:
\begin{equation}
   f(x)= 
   \begin{cases}
     x &\text{if } x \geq c \\
     s &\text{if } x<c
   \end{cases}
\end{equation}
with a cutoff $c$ and stable value $s$. In the experiments 
that follow we set $s=0$, returning maximum realness, and vary $c$ to test its effect. 

With the cut-off function, the search is unaffected until the image reaches a certain threshold of realness.  When the threshold is reached the system has no variation in respect to the realness value, thus giving priority to aesthetic features over realness values.

\begin{figure}[th]
\centering
\includegraphics[width=10.0cm]{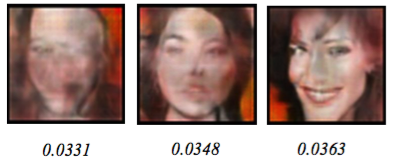}
\caption{The results of minimizing hue with cut offs at 0.2 (left), 0.05 (middle) and 0.02 (right).}
\label{first_cut_off_6}
\end{figure}

A subjective analysis of possible cut-off values needed to be performed in order to determine the optimal value for future experiments. 
Figure~\ref{first_cut_off_6} demonstrates the effect of different cut-off values on both image realness and aesthetic feature value.
A relationship can be observed from the above images. As the cut-off decreases, so does the degree we are able to evolve the feature value. However, it can be noted that even the $0.02$ cut-off was able to create a far lower hue compared to the constrained results -- while still being realistic enough to be called a face. The $0.02$ cut-off was used in the remaining single feature experiments. 

\section{Results}
\label{results}
This section presents the results of evolution in single and paired feature spaces. The previous experiments were all carried on faces. In the following \gan's generated from both the faces and the butterfly datasets were used.

\begin{figure}

	\centering

\subfloat[Hue] { 
\includegraphics[width=2.06cm]{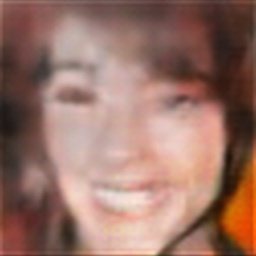}\label{pre:A}
\includegraphics[width=2.06cm]{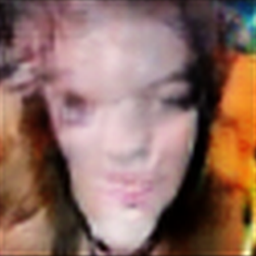}
\includegraphics[width=2.06cm]{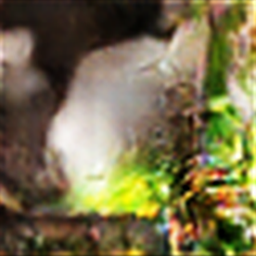}
\includegraphics[width=2.06cm]{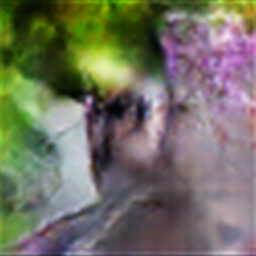}

}

\vspace{-0.25cm}
\subfloat[Saturation] {
\includegraphics[width=2.06cm]{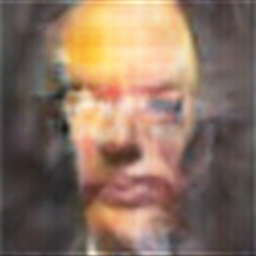}
\includegraphics[width=2.06cm]{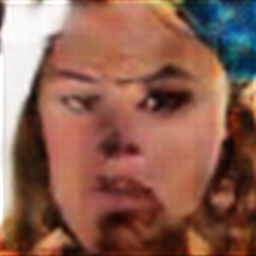}
\includegraphics[width=2.06cm]{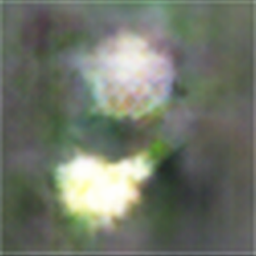}
\includegraphics[width=2.06cm]{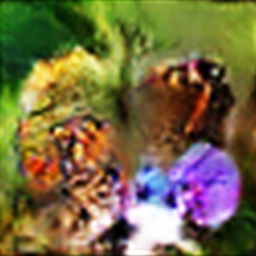}}

\vspace{-0.25cm}
\subfloat[Smoothness] {
\includegraphics[width=2.06cm]{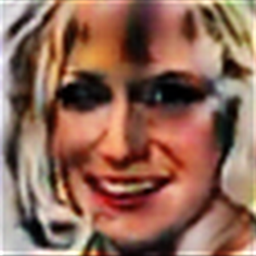}
\includegraphics[width=2.06cm]{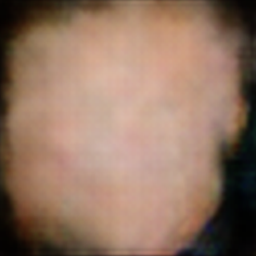}
\includegraphics[width=2.06cm]{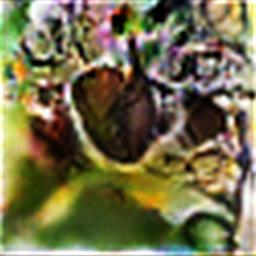}
\includegraphics[width=2.06cm]{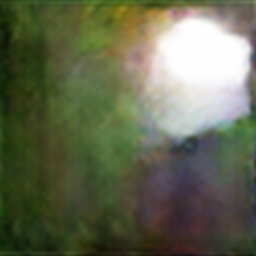}}

\vspace{-0.25cm}
\subfloat[Symmetry] {
\includegraphics[width=2.06cm]{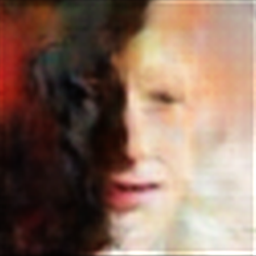}
\includegraphics[width=2.06cm]{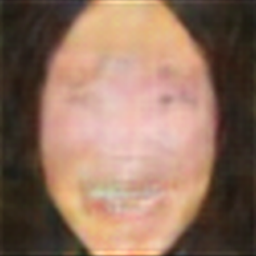}
\includegraphics[width=2.06cm]{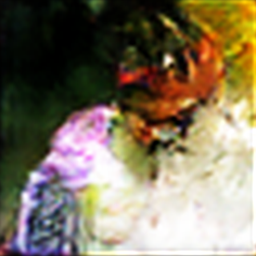}
\includegraphics[width=2.06cm]{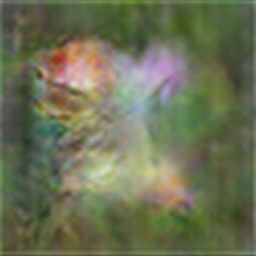}}

\vspace{-0.25cm}
\subfloat[GCF] {
\includegraphics[width=2.06cm]{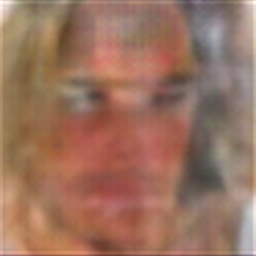}
\includegraphics[width=2.06cm]{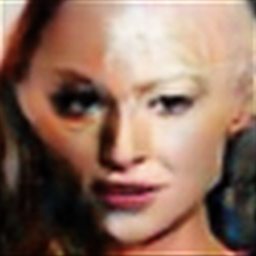}
\includegraphics[width=2.06cm]{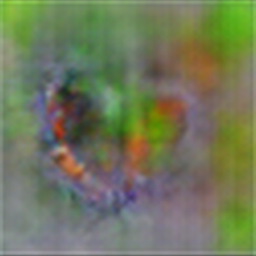}
\includegraphics[width=2.06cm]{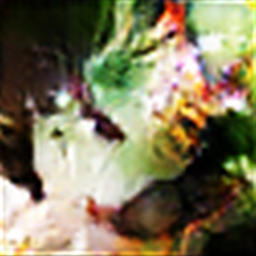}}

\caption{All single feature optimisations with 0.02 cut-off.}
\label{single_features_cut_off_7}
\end{figure}

\subsection{Single Dimensional Feature Experiments with Cut-off Function}
We conducted single feature dimension experiments using the face and butterfly datasets for the following features: hue, saturation, smoothness, symmetry and GCF. For these experiments we use a cut-off of $0.02$ on the discriminator output for both \gan's.  
The results were obtained for the minimum and maximum feature values from each experiment. Figure~\ref{single_features_cut_off_7} shows the result of the experiments
for the single dimensional feature with each row corresponding to an image minimising and maximising the feature for faces and butterflies respectively.

Table~\ref{tab:tab1} shows the minimum and maximum value for each feature for the faces and butterfly datasets, respectively. We observe that hue has the highest range with respect to feature values.
The use of the butterfly dataset provides good way to 
see how evolution with aesthetic measure responds to the priors embedded in different \gan's.

For the single dimensional experiments we observe in Figure ~\ref{single_features_cut_off_7} that images generated with the faces dataset appear more real than the images generated with butterfly dataset. This is likely to be due to the more diverse nature of the butterfly dataset. 
The images shown in Figure~\ref{single_features_cut_off_7} (a)
have the most variance in the hue dimension. Image with lowest hue value appears most realistic in contrast the image with higher value appears less realistic. We observe that image generated from the butterfly dataset achieves 
a higher feature range.

Figure~\ref{single_features_cut_off_7} (b) shows that in spite of the saturation feature for faces extending over a narrow range the resulting faces are not very realistic. The butterfly dataset is able to produce higher values of saturation, resulting in  a realistic and colorful image. 
In Figure~\ref{single_features_cut_off_7} (c) we observe that minimization for produces realistic images with superimposed darker shadows. In contrast maximising smoothness produces less realistic images. 

The images shown in Figure~\ref{single_features_cut_off_7} (d) produces high values for symmetry for both datasets. These images appears symmetrical and less real. Images with lower symmetry value are more real.

Finally, the images shown in Figure~\ref{single_features_cut_off_7} (e) exhibits less realness for faces and butterflies dataset in minimisation and more realism in maximisation.

\begin{table}
\centering
\begin{tabular}{l|r|r|r|r}
\hline

Feature  & Min-F & Max-F &Min-B&Max-B\\\hline
$\hue$ & 0.0337 & 0.4886 &0.1083&0.5282\\\hline
$Saturation$ & 0.3306 & 0.3543&0.1205&0.5918  \\\hline 
$Smoothness$ & 0.9737 & 0.9843 &0.9462&0.9887\\\hline
$Symmetry$ & 0.7985  & 0.9198 &0.5568&0.9428\\\hline
$\GCF$ &0.0106&0.0348&0.0090&0.0417 \\\hline 
\end{tabular}
\vspace{0.5 cm}
\caption{Single features value with cut-off of 0.2 for faces and butterfly dataset, from link respectively.}\label{tab:tab1}
\end{table}

\subsection{Two-Dimensional Feature Experiments with Cut-off Function}

In our next experiment, we evolve images using the $\gan$ to minimise and maximise in two feature dimensions. 
These experiments aim to give us insight into how features interact with each other and also the impact of the image priors as embedded in the \gan's on the extent to which features can be optimised. 
After training our $\gan$ models on the faces and butterfly dataset, 
We run experiments with the following feature combinations: GCF-Saturation; GCF-Smoothness; Hue-Saturation; Hue-Symmetry; Saturation-Symmetry; and Smoothness-Saturation. 

\begin{figure*}
\centering

\includegraphics[width=0.45cm]{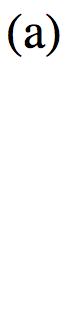}
\hspace*{0.2cm}
\includegraphics[width=3.76cm]{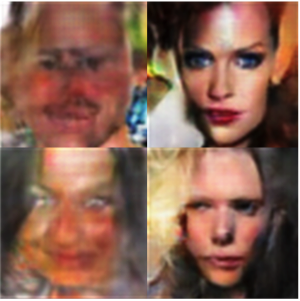}
\includegraphics[width=3.78cm]{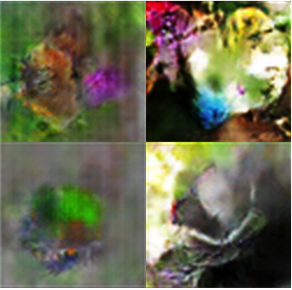}
\includegraphics[width=5.06cm]{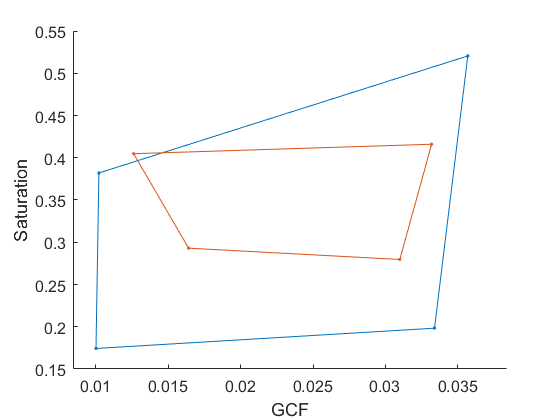}

\includegraphics[width=0.45cm]{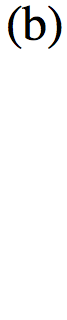}
\hspace*{0.2cm}
\includegraphics[width=3.76cm]{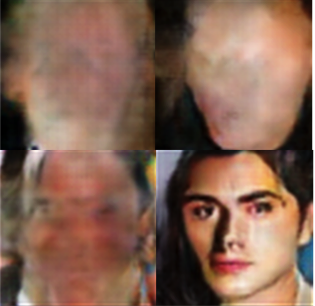}
\includegraphics[width=3.76cm]{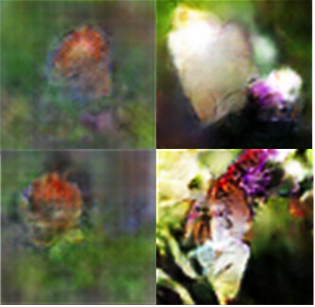}
\includegraphics[width=5.06cm]{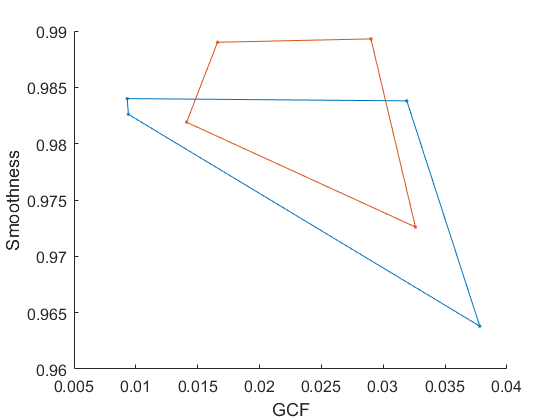}

\vspace{0.1cm}
\caption{Image obtained by multi-features with a 0.008 cut-off constraint. The images correspond to 2 feature pairs, GCF, saturation and GCF, smoothness for faces and butterfly dataset, from top left, respectively. Note the images follow their positions on the graph.}
\label{multi_features_cut_off_GCF_8}

\end{figure*}

\begin{figure*}
\begin{center}

\includegraphics[width=0.45cm]{a.png}
\hspace*{0.2cm}
\includegraphics[width=3.76cm]{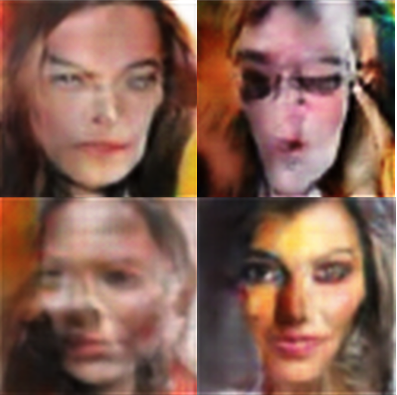}
\includegraphics[width=3.76cm]{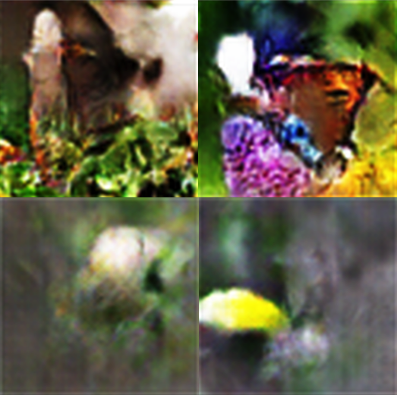}
\includegraphics[width=5.06cm]{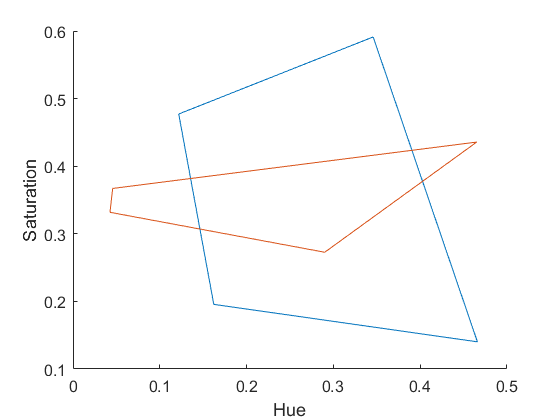}\\

\includegraphics[width=0.45cm]{b.png}
\hspace*{0.2cm}
\includegraphics[width=3.76cm]{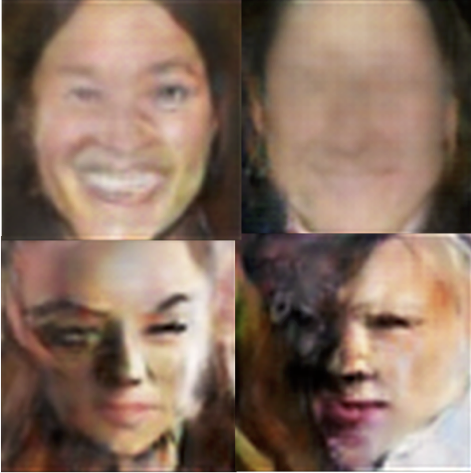}
\includegraphics[width=3.76cm]{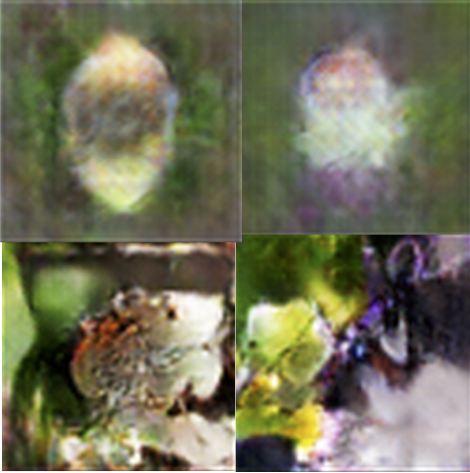}
\includegraphics[width=5.06cm]{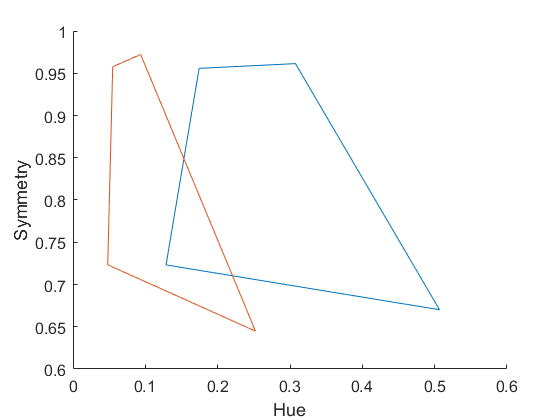}\\

\includegraphics[width=0.45cm]{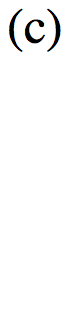}
\hspace*{0.2cm}
\includegraphics[width=3.76cm]{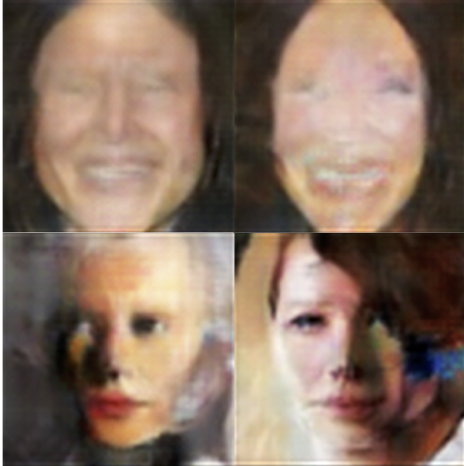}
\includegraphics[width=3.76cm]{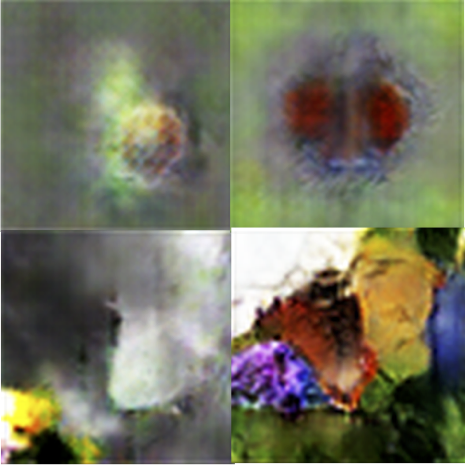}
\includegraphics[width=5.06cm]{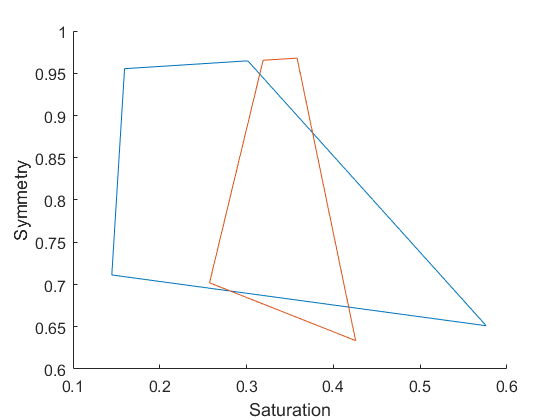}\\

\includegraphics[width=0.45cm]{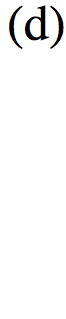}
\hspace*{0.2cm}
\includegraphics[width=3.76cm]{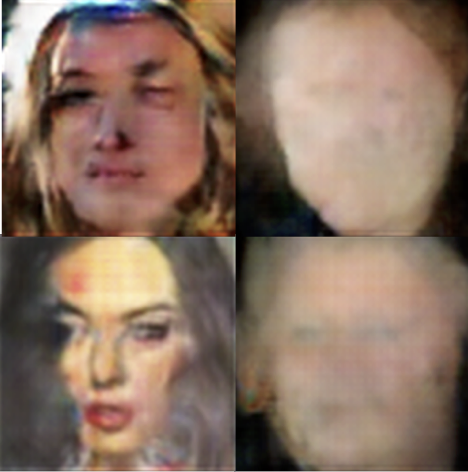}
\includegraphics[width=3.76cm]{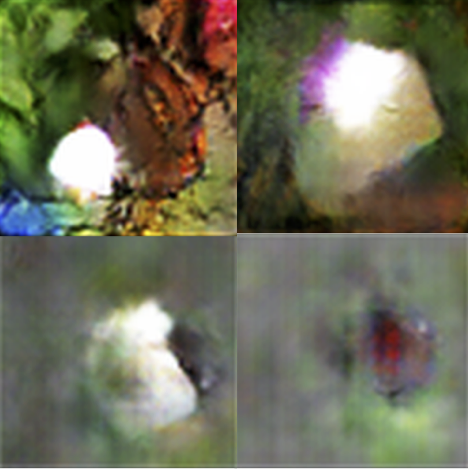}
\includegraphics[width=5.06cm]{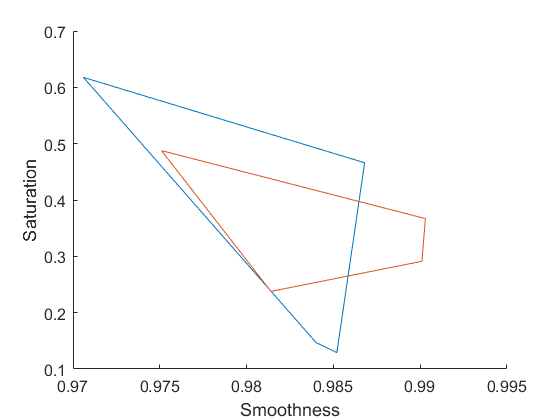}

\caption{Image obtained by multi-features with a 0.008 cut-off constraint. The first four images correspond to features hue and saturation, hue and symmetry, saturation and symmetry, smoothness and saturation, from top, respectively. The images follow their positions on the graph.}
\label{multi_features_cut_off_9}
\end{center}
\end{figure*}

The feature, pair values resulting from these experiments are shown in Tables~\ref{tab:tab1} (for faces) and~\ref{tab:tab2} (for butterflies). 
The images corresponding to these values are shown in Figure~\ref{multi_features_cut_off_GCF_8} and~\ref{multi_features_cut_off_9}. 
The first column of Figures\ref{multi_features_cut_off_GCF_8} and~\ref{multi_features_cut_off_9}, show  images, in clockwise order from top-left, for  min-max, max-max, max-min, min-min combinations of features for faces, 
The corresponding pictures butterflies dataset is shown in the second column. 
The last column, plots the positions in feature-space, of the four face images from the first column (in red) and the four butterfly images from the second column (in blue). 
The shape of the quadrilateral in these plots provides an 
indication of how feature values are constrained with respect to each other and by the $\gan$ used to generate them.

Based on our findings from the previous experiments with single dimension diversity, we reduced the cut-off to $0.008$ for the multi-feature experiments to try to maintain the realism of the images. The impact of this smaller cut-off can be observed in Figures~\ref{multi_features_cut_off_GCF_8} and ~\ref{multi_features_cut_off_9} in terms of the relatively small areas of feature space contained by the plots.

Looking at the feature-pairs in turn. 
The images shown in Figure~\ref{multi_features_cut_off_GCF_8} (a) have the highest GCF values and for maxmin and maxmax optimisation appear most realistic. We observe in Figure~\ref{multi_features_cut_off_GCF_8} (b) that image generated on the butterflies dataset achieve higher score for GCF and permit a higher range of saturation for high-GCF. 

The feature plot in Figure~\ref{multi_features_cut_off_GCF_8}
shows that GCF and Saturation can vary independently. 
In contrast, Figure~\ref{multi_features_cut_off_GCF_8} (b)'s plot indicates some difficulty in minimising both smoothness and GCF. From the plot also appears to be relatively difficult to maximise both of these features. 
This result is in concordance with the observations in ~\cite{DBLP:conf/gecco/AlexanderKN17} which found that GCF and smoothness, being spatial features, appeared to be in conflict with each other. More specifically the high contrast required for high GCF scores is in direct conflict with the low contrast required for high smoothness scores. Also notable from figure~\ref{multi_features_cut_off_GCF_8} (b) is a relative lack of realism in the faces as compared to those in part (a).

The images shown in Figure~\ref{multi_features_cut_off_9} (a) 
show the relationship between hue and saturation. The butterfly pictures show the most variance in the saturation dimension and the face pictures show marginally more variance in hue. Images high in saturation seem to appear sharper -- with the face image that maximises both features having quite harsh colour, more contrast, and a mask-like appearance.

For figure~\ref{multi_features_cut_off_9} (b) both sets of images have similar ranges of symmetry but faces have a much narrower range of hue. Highly symmetric images seem to be less realistic, tending to ovoid shapes, with detail seemly sacrificed in order to maximise symmetry.  In contrast asymmetric images appear to have more realistic textures and more intense colours.

Figure~\ref{multi_features_cut_off_9} (c) combines saturation and symmetry. As before highly symmetric images appear less realistic. The evolutionary process seems to have difficulty maximising both saturation and symmetry for both \gan's. Clearly it is possible to create artificial images that score highly on both feature dimensions so this difficulty may be reflective of the rarity of this feature combination in the training sets for these \gan's. As a final observation, the butterfly picture maximising both features resembles an insect's face, perhaps an interesting consequence of having diverse images in the training set.

Finally, the images in Figure~\ref{multi_features_cut_off_9} (d) 
show difficulty in minimising both smoothness and saturation.  
There is a smaller corresponding problem in maximising both features. In both data sets the most realistic images are produced by the minimisation of smoothness and the maximisation of saturation, perhaps indicating that the priors in the data set are biased toward rougher and more colourful images.

\begin{table*}
\centering
\begin{tabular}{l|r|r|r|r}
\hline

Feature  pairs &  Min.f1-Min.f2  &  Min.f1-Max.f2  &  Max.f1-Min.f2  & Max.f1-Max.f2 \\\hline
$Hue-Sat.$ & 0.0426 - 0.3318 & 0.0457 - 0.3671 & 0.2900 - 0.2727 & 0.4651 - 0.4257 \\\hline
$Hue-Sym.$ & 0.0480 - 0.7233 & 0.0549 - 0.9577& 0.2523 - 0.6448 & 0.0935 - 0.9722 \\\hline 
$Sat.-Sym.$ & 0.2573 - 0.7020 & 0.3190 - 0.9654 & 0.4256 - 0.6336 & 0.3582 - 0.9679 \\\hline
$Smooth.-Sat.$ & 0.9814 - 0.2378 & 0.9751 - 0.4876 & 0.9901 - 0.2912 & 0.9903 - 0.3670 \\\hline
$\GCF-Sat. $ & 0.0164 - 0.2930 & 0.0126 - 0.4048 & 0.0310 - 0.2796 & 0.0332 - 0.4160 \\\hline 
$\GCF-Smooth.$ & 0.0166 - 0.9890 & 0.0166 - 0.9890 & 0.0326 - 0.9762& 0.0290 - 0.9893 \\\hline 
\end{tabular}
\caption{Dual features with cut-off for faces dataset.}\label{tab:tab1}
\vspace{3mm}
\begin{tabular}{l|r|r|r|r}
\hline
Feature pairs & Min.f1-Min.f2 & Min.f1-Max.f2 & Max.f1-Min.f2 & Max.f1-Max.f2 \\\hline
$Hue-Sat.$ & 0.1622 - 0.1956 & 0.1218 - 0.4772 & 0.4659 - 0.1402 & 0.3458 - 0.5912 \\\hline
$Hue-Sym.$ & 0.1286 - 0.7232 & 0.1744 - 0.9558& 0.5067 - 0.6701& 0.3075 - 0.9614 \\\hline 
$Sat.-Sym.$ & 0.1447 - 0.7133 & 0.1594 - 0.9554 & 0.5760 - 0.6512 & 0.3014 - 0.9647\\\hline
$Smooth.-Sat.$ & 0.9840 - 0.1469 & 0.9706 - 0.6177 & 0.9852 - 0.1291 & 0.9868 - 0.4661 \\\hline
$\GCF-Sat.$ & 0.0100 - 0.1743 & 0.0102 - 0.3820 & 0.0334 - 0.1983 & 0.0357 - 0.5205 \\\hline 
$\GCF-Smooth.$ & 0.0094 - 0.9826 & 0.0093 - 0.9840 & 0.0378 - 0.9638 & 0.0319 - 0.9838 \\\hline 
\end{tabular}
\caption{Dual features with cut-off for butterfly dataset.}\label{tab:tab2}
\end{table*}

\section{Discussion and future work}
\label{discussion}

 Evolutionary search can be a powerful technique for creating novel images. 
In this work, we have shown how to apply \gan's in order to generate images scoring high or low for given aesthetic feature values. We used evolutionary search to maximise and minimise single features and pairs of features for two datasets, faces and butterflies. 

We have shown how to explore the latent space of a ${\gan}$ to create semi-realistic images that sample different regions of aesthetic feature spaces. 
Additionally, we studied the effects of different values of the cut-off function on the aesthetic appearance of the images.

For future research, it would be interesting to explore
intermediate points in the feature space to gain more insight into the relationships between features and
to explore additional constraints and their effect on the process of generating novel images.



\end{document}